%% file: acl_latex.tex
\pdfoutput=1

\documentclass[11pt]{article}

\usepackage[preprint]{acl}

\usepackage{times}
\usepackage{latexsym}

\usepackage[T1]{fontenc}

\usepackage[utf8]{inputenc}

\usepackage{microtype}

\usepackage{inconsolata}

\usepackage{graphicx}
\usepackage{latexsym}
\usepackage{makecell}
\usepackage{amsmath,amssymb,bm}
\usepackage{url}
\usepackage{diagbox}
\usepackage[ruled, linesnumbered, titlenumbered]{algorithm2e}
\usepackage{bbm}
\usepackage{multirow}
\usepackage{booktabs}
\usepackage{pifont}
\usepackage{siunitx}
\usepackage{array}
\usepackage{colortbl}
\usepackage{ulem}
\usepackage{pgfplots}
\usepackage{dashrule}
\usepackage{makecell}
\usepackage{subcaption}

\title{Revisiting Compositional Generalization Capability of Large Language Models Considering Instruction Following Ability}

\author{Yusuke Sakai, \
        Hidetaka Kamigaito, \
        Taro Watanabe \\
  Nara Institute of Science and Technology (NAIST), Japan  \\
  \texttt{\{sakai.yusuke.sr9, kamigaito.h, taro\}@is.naist.jp}}

\begin{document}
\maketitle
\begin{abstract}

In generative commonsense reasoning tasks such as CommonGen, generative large language models (LLMs) compose sentences that include all given concepts. However, when focusing on instruction-following capabilities, if a prompt specifies a concept order, LLMs must generate sentences that adhere to the specified order. To address this, we propose Ordered CommonGen, a benchmark designed to evaluate the compositional generalization and instruction-following abilities of LLMs. This benchmark measures ordered coverage to assess whether concepts are generated in the specified order, enabling a simultaneous evaluation of both abilities. We conducted a comprehensive analysis using 36 LLMs and found that, while LLMs generally understand the intent of instructions, biases toward specific concept order patterns often lead to low-diversity outputs or identical results even when the concept order is altered. Moreover, even the most instruction-compliant LLM achieved only about 75\% ordered coverage, highlighting the need for improvements in both instruction-following and compositional generalization capabilities.

\end{abstract}

\section{Introduction}

\begin{figure}[t]
\centering
  \includegraphics[width=\columnwidth]{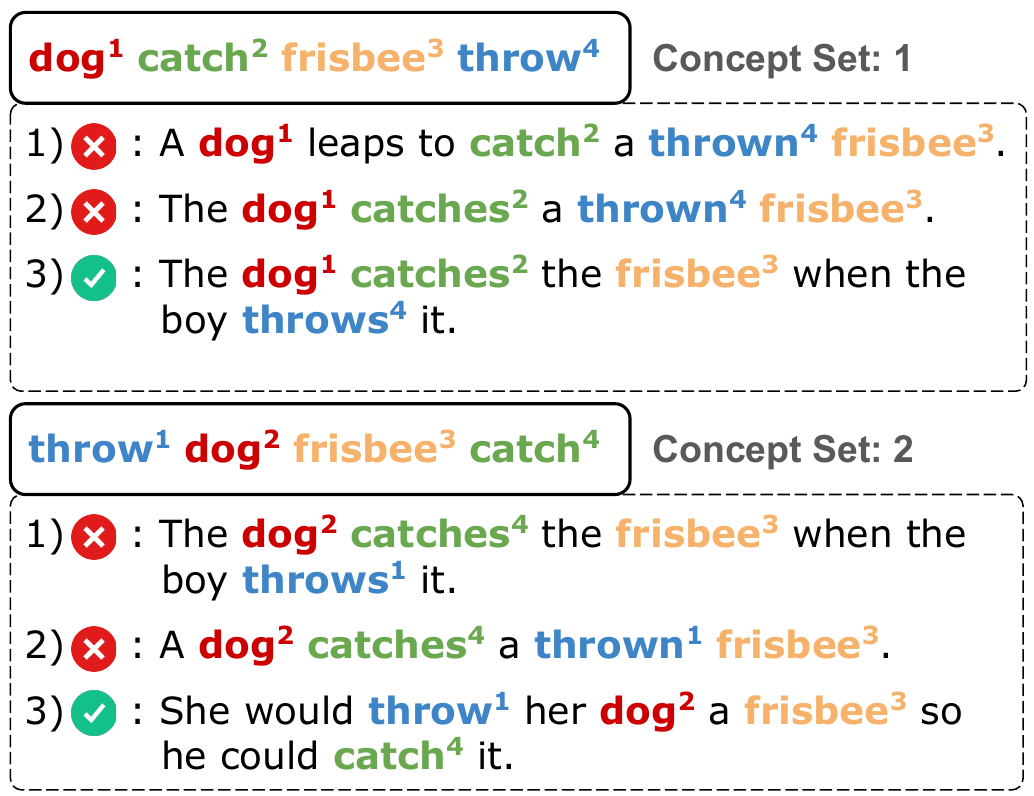}
  \caption{Overview of our proposed Ordered CommonGen. Unlike CommonGen~\cite{lin-etal-2020-commongen}, we evaluate whether the composed sentences include the concepts in the specified order. To create the Ordered Concept Sets, we use CommonGen’s Concept Sets containing four concepts and generate all permutations, resulting in a total of 24 permutations per set.}
  \label{fig:overview}
\end{figure}

\begin{figure*}[!t]
\centering
  \includegraphics[width=0.95\linewidth]{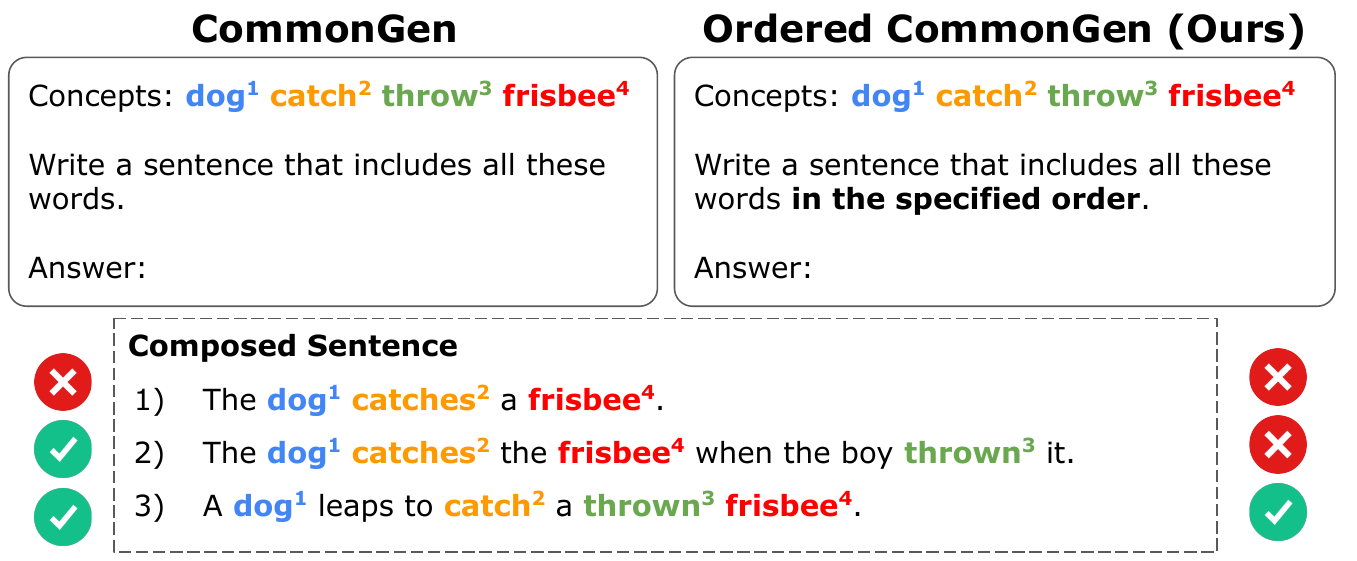}
  \caption{Our evaluation methodology for Ordered CommonGen. The left side shows the instruction template for the standard CommonGen task, while the right side shows the template used in our Ordered CommonGen. All templates are provided in Appendix~\ref{sec:example-evaluation-templates}. In Ordered CommonGen, the phrase \textbf{``in the specified order''} is inserted to explicitly instruct the LLM to compose sentences following the given concept order. For example, Given a concept set (dog, catch, throw, frisbee), the composed sentence (1) does not include all the concepts, making it incorrect for both tasks; (2) includes all the concepts but involves reordering, making it correct only for CommonGen; and (3) includes all the concepts in the specified order, making it correct for both tasks.}
  \label{fig:template-example}
\end{figure*}

With the maturity of instruction-tuning techniques such as supervised fine-tuning~\cite{wei2022finetuned, sanh2022multitask, ghosal2023flacunaunleashingproblemsolving, wang-etal-2023-self-instruct} and human-preference tuning~\cite{ziegler2020finetuninglanguagemodelshuman, NEURIPS2022_b1efde53_gpt3.5, winata2024preferencetuninghumanfeedback}, generative large language models (LLMs) are capable of producing high-quality, fluent responses aligned with user instructions. 
Leveraging their instruction-following capabilities and advanced text generation abilities, LLMs are applied not only to general downstream tasks such as dialogue systems~\cite{wang2023surveyevolutionlanguagemodelbased, yi2024surveyrecentadvancesllmbased, li2024helloagainllmpoweredpersonalized, zhao2024wildchat} but also to creative domains, including dataset construction through complex prompts tailored to specific purposes~\cite{tan-etal-2024-large, sakai-etal-2024-simultaneous, makinae-etal-2024-simul}, as well as crafting catchy slogans~\cite{kim-slogan2023, brigham2024developingstorycasestudies}, advertisements~\cite{lei-etal-2022-plato, mita-etal-2024-striking}, and poetry~\cite{yu-etal-2024-charpoet, zhang2024llmbasedmultiagentpoetrygeneration}.
However, the instruction-following abilities of LLMs are still developing, and they sometimes produce outputs that deviate from intended instructions~\cite{palmeira-ferraz-etal-2024-llm, qian-etal-2024-tell}. This highlights the need to benchmark their ability to consistently generate text that strictly adheres to given instructions.

Generative commonsense reasoning (GCR)~\cite{lin-etal-2020-commongen, nan-etal-2021-dart, seo-etal-2022-dog, liu-etal-2023-dimongen} is a type of constrained text generation task that requires compositional generalization capabilities, where LLMs are tasked with creating natural, commonsense sentences that include all given concepts in Figure~\ref{fig:overview}. Traditional GCR tasks evaluate coverage, focusing on whether the given concepts are included in the composed sentences without considering their order. Consequently, LLMs often rearrange the order of concepts to produce more natural sentences~\cite{ou-etal-2022-role, zhang-etal-2023-learning-predict}. 
However, when instructed to follow a specific order, LLMs must compose sentences that adhere to the user-specified order of concepts.

Such potential needs are particularly important in creative domains, such as describing events in chronological order~\cite{mostafazadeh-etal-2016-corpus, lu-etal-2023-narrative, Chen2023follow, pawlowski-walkowiak-2024-nlp}, planning actions~\cite{sakib2023cookingrecipesrobottask, lin2024graphenhancedlargelanguagemodels, singh-etal-2024-personal}, or composing lyrics~\cite{fan2019hierarchical, hu-sun-2020-generating, qian-etal-2023-unilg}, where a change in the order of concepts can significantly alter the meaning or nuance of the output. 
According to generative grammar~\cite{chomsky1969aspects, Jackendoff2002}, natural sentences can be composed following syntactic rules without arbitrary rearrangements, as humans are capable of processing and generating sentences in grammatically permissible orders.
Therefore, by simultaneously addressing instruction-following ability and compositional generalization ability, LLMs can be applied to more creative text generation tasks.

To address this, we propose Ordered CommonGen, a framework designed to evaluate both compositional generalization and instruction-following abilities by reorganizing CommonGen~\cite{lin-etal-2020-commongen}, a representative GCR task. As shown in Figure~\ref{fig:overview}, Ordered CommonGen permutes the order of concepts and, as illustrated in the prompts in Figure~\ref{fig:template-example}, inserts the phrase \textbf{``in the specified order''} into the instruction. This setup enables the evaluation of whether LLMs can compose sentences while adhering to the given instructions.

Through a comprehensive benchmarking across 36 LLMs, we found that while LLMs can compose sentences containing the given concepts, their ability to compose sentences in the specified order, as required by Ordered CommonGen, is limited to about 75\%, even for the best-performing model. Interestingly, inserting the phrase \textbf{``in the specified order''} into the instruction improves the ability to follow the specified order, indicating that LLMs understand the intent of the instructions. However, biases toward specific concept order patterns and identical outputs, even when the order of concepts is altered, remain significant challenges. These findings highlight challenges in the compositional generalization and instruction-following capabilities of LLMs within GCR tasks.

\section{Ordered CommonGen}

\subsection{Task Definition}

Generative Commonsense Reasoning (GCR) tasks such as CommonGen~\cite{lin-etal-2020-commongen} involve generating a natural and grammatically correct sentence that incorporates all given concepts in a concept set. Formally, the input consists of a concept set $X = \{c_1, c_2, \dots, c_k\}$, where each concept $c_i$ is a lemmatized common object (noun) or action (verb). The task requires LLMs to generate a sentence $Y$ such that all concepts in $X$ are included, allowing for morphological inflections or variations. The inclusion of concepts is judged by verifying if the lemmatized forms of all concepts in $X$ are present in $Y$. Formally, the output is represented as $Y = \{y_1, y_2, \dots, y_n\}$, where $y_i$ is the $i$-th sentence generated given the input $X$. A generated sentence $Y$ is considered valid if it contains all concepts in $X$ in any order, satisfying $X \subseteq Y$.

In Ordered CommonGen, beyond the traditional requirements of GCR tasks, we introduce an additional dimension to evaluate the instruction-following ability of LLMs. Specifically, we assess whether the generated sentence adheres to the \textbf{specified order of the concepts} in the input concept set $X$, checking them sequentially from the beginning.

Humans, capable of the \textit{``infinite use of finite means''}~\cite{chomsky1969aspects}, can compose sentences following any specified order of concepts, regardless of the arrangement of the words themselves~\cite{Bever1970, Levelt1989, FERREIRA200661, goldberg1995constructions, Jackendoff2002}. %
This expanded evaluation enables us to explore both compositional generalization, which refers to the systematic combination of known elements in novel ways~\cite{Lake2018, lake2018generalizationsystematicitycompositionalskills}, and instruction-following abilities, which reflect the capacity to align generated outputs with explicit external directives~\cite{Lupyan2015}.
This evaluation highlights the gap between human linguistic understanding and the ability of LLMs.

\subsection{Dataset Construction}

As shown in Figure~\ref{fig:template-example}, two key components are required for dataset creation: \textbf{concept sets}, which are collections of word lists used as seeds for text generation, and \textbf{instruction templates}, which incorporate the concept sets and serve as input prompts for LLMs.
First, we used CommonGen-lite\footnote{\url{https://hf.co/datasets/allenai/commongen_lite}}, a subset of the CommonGen test data, as a seed dataset. We extracted all 192 seed concept sets, each consisting of four concepts. By generating all possible permutations (4! = 24) of the concepts in each case, we created a total of 192 $\times$ 4! = 4,608 concept sets.

Next, we curated the instruction templates from FLAN~\cite{wei2022finetuned, longpre2023flancollectiondesigningdata}, a collection of manually created templates for various task datasets. Specifically, we utilized all six instruction templates designated for CommonGen in FLAN as base templates. By averaging scores across the six multi-templates, our evaluation accounts for variance in template performance~\cite{sakai-etal-2024-toward}.
We modified the base instruction templates by appending the phrase \textbf{``in the specified order''}. This template requires LLMs to generate sentences that adhere to the order of the concept set.
All examples of instruction templates are provided in Appendix~\ref{sec:example-evaluation-templates}.
Finally, we obtained a total of 6 $\times$ 4,608 = 27,648 instances.

\section{Experimental Settings}

\subsection{Evaluation Metrics}

\paragraph{Concepts Coverage:}
We report three types of coverage rates.
First, following~\citet{lin-etal-2020-commongen}\footnote{We referred to the implementation at \url{https://github.com/allenai/CommonGen-Eval}. For lemmatizing the generated sentences, we similarly used spaCy~\cite{spacy} with the \textsc{en\_core\_web\_lg} model: \url{https://spacy.io/models/en\#en_core_web_lg}.}, we report the average percentage of input concepts included in the lemmatized outputs, disregarding their order (\textbf{Coverage w/o order}).
Next, aligned with the aim of Ordered CommonGen, we evaluate whether the generated sentences include all input concepts in the specified order (\textbf{Coverage w/ order}).
Finally, we report the average percentage of generated sentences that include all input concepts and adhere to the specified order (\textbf{Ordered Rate}).

\paragraph{Sentence-level Similarity:}
When the given order of concepts varies, it influences the structure and coherence of composed sentences, as certain syntactic and semantic constraints govern how concepts can be naturally combined~\cite{goldberg1995constructions, Jackendoff2002, Steedman2000}. Therefore, it is important to produce diverse outputs that adhere to different orders of the concept set.
Following the idea of Pairwise-BLEU~\cite{pmlr-v97-shen19c}, we evaluate the diversity of generated sentences by computing the average pairwise similarity across all 24 sentences obtained from permutations of the same four concepts. A lower similarity score suggests greater diversity, as it indicates less overlap among generated sentences.
To assess similarity at different levels, we employ two metrics: Pairwise-BLEU (\textbf{pBLEU}), which measures surface-level $n$-gram overlap using BLEU~\cite{papineni-etal-2002-bleu, post-2018-call}, and Pairwise-BLEURT (\textbf{pBLEURT}), which evaluates semantic similarity using BLEURT~\cite{sellam-etal-2020-bleurt}. %
For BLEU, we set $n=4$, and for BLEURT, we used the \textsc{BLEURT-20} model.

\paragraph{Corpus-level Diversity:}

We also evaluate the overall diversity of the generated sentences using distinct-$n$~\cite{li-etal-2016-diversity}\footnote{\url{https://github.com/neural-dialogue-metrics/Distinct-N}} (\textbf{Distinct}), which is calculated as the ratio of unique $n$-grams to the total number of $n$-grams in all texts formulated as:
\begin{equation}
    \text{distinct-}n = \frac{\#\text{unique } n\text{-grams}}{\#\text{total } n\text{-grams}}.
\end{equation}
We set $n=2$.
Furthermore, we observed in our preliminary study that some outputs remain identical despite the different order of concepts. This occurs because LLMs have the ability to rearrange concepts to generate more natural sentences~\cite{ou-etal-2022-role, zhang-etal-2023-learning-predict}. However, from the perspective of instruction-following ability, this behavior is inappropriate. Therefore, we introduce a new diversity metric, \textbf{Diverse Rate}, to quantify variations in generated sentences.
The Diverse Rate is calculated as the ratio of unique sentences to the total number of sentences formulated as: 
\begin{equation}
    \text{Diverse Rate} = \frac{\#\text{unique sentences}}{\#\text{total sentences}}.
\end{equation}
If all generated sentences are unique, the Diverse Rate is 1.00. For instance, if 24 sentences are generated but only one is unique, the Diverse Rate would be $ \frac{1}{24} \fallingdotseq 0.04$.
As these metrics represent the ratio of unique $n$-grams or sentences, a higher score indicates greater diversity.

\paragraph{Perplexity:}

In GCR tasks such as CommonGen, the quality of generated sentences is typically evaluated by comparing them to human references. However, our focus is on generating diverse sentences, and the references do not account for all permutations of the concept sets. Therefore, these evaluation methods are not aligned with our goal. Instead, we employ \textsc{GPT2-XL}~\cite{radford2019language}\footnote{\url{https://hf.co/openai-community/gpt2-xl}} as an evaluation model and measure perplexity via \textsc{lmppl}\footnote{\url{https://github.com/asahi417/lmppl}} to perform a relative assessment of the quality of the generated sentences. Since perplexity is equivalent to the information content of a sentence, lower perplexity generally indicates higher-quality sentences. Furthermore, LLMs capable of generating sentences with lower perplexity while adhering to the specified order are more likely to produce commonsensical and coherent sentences.

\subsection{Evaluation Setup}
\paragraph{Settings for LLMs.}
We primarily selected a total of 36 well-known instruction-tuned LLMs for evaluation: Llama3-3.2-1B, 3.2-3B, 3.1-8B, 3.1-70B, 3.3-70B, 3.1-405B~\cite{dubey2024llama3herdmodels}, Qwen2-0.5B, 1.5B, 7B, 72B~\cite{yang2024qwen2technicalreport}, Qwen2.5-0.5B, 1.5B, 3B, 7B, 14B, 32B, 72B~\cite{qwen2.5}, Gemma2-2B, 9B, 27B~\cite{gemmateam2024gemma2improvingopen}, Phi3-mini, small, medium~\cite{abdin2024phi3technicalreporthighly}, T\"ulu3-8B, 70B~\cite{lambert2024tulu3pushingfrontiers}, OLMo2-7B, 13B~\cite{olmo2}, Mistral-7B, Small, Large~\cite{jiang2023mistral7b}, Mixtral-8x7B, 8x22B~\cite{jiang2024mixtralexperts}, Gemini-Flash, Pro~\cite{geminiteam2024gemini15unlockingmultimodal}, GPT-3.5~\cite{NEURIPS2022_b1efde53_gpt3.5}, and GPT-4o~\cite{openai2024gpt4technicalreport}. Following the evaluation settings by~\citet{sakai-etal-2024-mcsqa}, for proprietary models (Gemini, GPT-3.5, and GPT-4o), the temperature was set to 0 to achieve as deterministic outputs as possible. For the other open LLMs, outputs were generated using greedy decoding with a fixed seed value and 4-bit quantization~\cite{dettmers2023qlora}\footnote{The 4-bit quantization models yield optimal performance~\cite{dettmers2023qlora, liu-etal-2024-emergent, dettmers2023case}. Therefore, we apply 4-bit quantization to save computational costs.}. 
Further details are provided in Appendix~\ref{sec:detailed-llm-information}.

\paragraph{Settings for Datasets.}
We compared two instruction prompts: one using our Ordered CommonGen template with the phrase \textbf{``in the specified order''}, and the other using the original CommonGen template without this phrase. If including the phrase in the prompt improves order-considered coverage, it indicates that the model successfully follows the given instructions. We report average scores across all six templates, with Ordered CommonGen values shown alongside their differences from the original CommonGen.%
We primarily conducted evaluations in zero-shot settings to highlight differences in inductive reasoning ability.

\section{Results and Discussions}

\input{main_result}

Table~\ref{tab:main_results} shows the main evaluation results.

\paragraph{Finding 1: LLMs understand the intent of instruction prompts.}

Focusing on the increase in the w/ order scores in Table~\ref{tab:main_results}, specifying the usage order of concepts in the prompt through Ordered CommonGen improves coverage rates for most LLMs. This is further supported by the rise in Ordered Rate scores. Notably, for certain LLMs such as Llama3.3-70B, Llama3.1-405B, and GPT-4o, w/o order scores remain high while w/ order scores increase significantly, indicating that these models attempt to align with the specified order. This suggests that LLMs strive to produce outputs consistent with the intent of instruction prompts.
Interestingly, even in the w/o order scores, Ordered CommonGen improves coverage rates for most LLMs. This suggests that explicit instructions to include concepts in the output encouraged LLMs to use them effectively, even when they struggled to arrange them in the specified order.

\paragraph{Finding 2: LLMs tend to generate natural sentences while trying to follow the instructions.}
When focusing on perplexity in Table~\ref{tab:main_results}, specifying the order of concepts in the input prompt generally worsens perplexity compared to when no order is specified. However, the degradation is relatively limited, except for certain LLMs such as OLMo2-13B and GPT-3.5. This indicates that LLMs are capable of following instructions while still generating natural sentences. Furthermore, LLMs respond to the phrase ``in the specified order'' included in Ordered CommonGen, demonstrating their ability to switch generation behavior accordingly while maintaining compositional generalization abilities. 
As the increase in perplexity is generally minimal, yet the Ordered Rate improves, this suggests that LLMs can generate sentences that adhere to the specified order while maintaining a degree of naturalness.
It suggests behavior resembling semantic compositionality~\cite{Jackendoff2002}.

\paragraph{Finding 3: Yet LLMs often struggle to generate sentences that follow instructions precisely.}
On the other hand, even the LLM with the highest Ordered Rate, Llama3.1-405B, followed instructions in only about 75\% of cases, leaving over 20\% of outputs that did not adhere to the given instructions. Furthermore, w/o order scores did not reach 100\%, indicating that LLMs fail to fully comply with instructions. For example, in the case of Llama3.1-405B, given the concept set (\textit{number, wear, shirt, run}), the generated sentence \textit{``The large \textbf{number} on the back of my \textbf{shirt} seemed to motivate me to \textbf{run} even faster.''} excludes the concept \textit{wear}. In contrast, humans would compose sentences such as \textit{``A large \textbf{number} of people \textbf{wear} a yellow \textbf{shirt} to \textbf{run} in the charity marathon.''}, naturally including all concepts in the specified order. Such omissions of concepts highlight a critical challenge in GCR tasks~\cite{lin-etal-2020-commongen, zhao-etal-2022-revisiting, zhang-etal-2023-learning-predict}. Moreover, the additional constraint of maintaining the specified order in Ordered CommonGen makes this task even more challenging.%
We provide a more detailed analysis of such cases in Appendix~\ref{sec:case-study}, where we show, based on statistical evidence, that at least one of the 36 LLMs we tested is capable of composing each sentence correctly.
This reflects the fact that many LLMs still have some room in their compositional capabilities.

\paragraph{Finding 4: LLMs generate more diverse outputs when concept order is considered.}

Focusing on Sentence-level Similarity in Table~\ref{tab:main_results}, LLMs such as Llama3.1-405B, Qwen2.5-72B, and GPT-4o, which show significant improvements in Ordered Rate through Ordered CommonGen, also demonstrate substantial gains in both pBLEU and pBLEURT. In contrast, some LLMs, such as GPT-3.5 and Gemini-Flash, show improvements in pBLEU but a decline in pBLEURT. 
Interestingly, in some LLMs, such as Qwen, smaller models tend to generate more diverse outputs than their larger ones at the expense of Concepts Coverage.
This suggests that while adhering to instructions improves syntactic diversity, it does not necessarily enhance semantic diversity. These results indicate that LLMs prioritize syntactic compositionality, as reflected in consistent pBLEU improvements~\cite{goldberg1995constructions, Steedman2000, lake2018generalizationsystematicitycompositionalskills}, while semantic compositionality remains more challenging due to its reliance on deeper contextual understanding~\cite{Jackendoff2002, bresnan2001lfs, bender-koller-2020-climbing}. This suggests that LLMs with higher instruction-following abilities, such as Llama3.1-405B, better balance syntactic and semantic compositionality, whereas lower-performing LLMs rely more heavily on syntactic patterns.

\paragraph{Finding 5: LLMs sometimes generate identical sentences even when the concepts are shuffled.}
According to the Diverse Rate in Table~\ref{tab:main_results}, even the highest-performing LLM, Mixtral-8x7B, does not reach 100\%, indicating that identical sentences are sometimes generated despite changes in the order of concepts. Furthermore, LLMs such as Gemma, Gemini, and OLMo2 achieve only around 80\%, suggesting a tendency to reorder concepts into sequences they consider most natural, overriding instructions to follow the specified order. 
For example, in the case of Gemma2-2B with the concept set (\textit{dock, dog, jump, water}), the output remains the same as \textit{\textbf{``The dog jumped off the dock into the water.''}} across all permutations of the input concepts for a given prompt.
This behavior reflects the influence of frequent concept set orders in the training data~\cite{zhang-etal-2023-learning-predict, ou-etal-2022-role, zhao-etal-2022-revisiting, yang-etal-2023-bridging, bender-koller-2020-climbing}, which often take precedence over given instructions. 
Interestingly, models like GPT-4o and Llama3.3-70B tend to produce duplicate outputs when instructions do not enforce a specified order. In such cases, LLMs default to generating sentences in the most natural sequence, regardless of input order.
This tendency to default to natural sequences aligns with usage-based theories~\cite{Bybee+2006, bybee2010language}, which argue that frequent patterns shape human language production, establishing the `default' or `natural' order.
Similar tendencies are observed in downstream tasks like neural machine translation~\cite{raunak-menezes-2022-finding} and grammatical error correction~\cite{cao-etal-2023-unnatural}, where minor input variations are absorbed, and frequent patterns influence outputs in the training data. Addressing these limitations may require improved training methods or increased model parameters. Additionally, well-instruction-following LLMs like GPT-4o and Llama3.1-405B demonstrate the ability to switch behaviors based on the instructions.

\section{Analysis}

\begin{table}[!t]
\centering
\resizebox{\linewidth}{!}{%
\small
\setlength{\tabcolsep}{2.2pt}
\rowcolors{2}{gray!10}{white}
\begin{tabular}{@{}cllll@{}}
\toprule
\rowcolor{white}
POS & \multicolumn{3}{c}{Concepts Coverage} & \multicolumn{1}{c}{Diverse} \\ \cmidrule(){2-4}
\rowcolor{white}
pattern & \multicolumn{1}{c}{w/o order} & \multicolumn{1}{c}{w/ order} & \multicolumn{1}{c}{Ordered Rate} &  \multicolumn{1}{c}{Rate}\\
\midrule
NNNN & $\textbf{91.13}_{\pm11.86}$ & $40.92_{\pm15.70}$ & $44.88_{\pm16.39}$ & $91.97_{\pm6.78}$ \\
NNNV & $\underline{88.41}_{\pm10.70}$ & $29.39_{\pm14.26}$ & $33.04_{\pm14.57}$ & $86.94_{\pm8.57}$ \\
NNVN & $88.06_{\pm10.46}$ & $37.34_{\pm13.69}$ & $42.09_{\pm13.40}$ & $86.94_{\pm8.57}$ \\
NNVV & $84.26_{\pm12.61}$ & $27.45_{\pm14.55}$ & $32.34_{\pm15.69}$ & $89.58_{\pm6.81}$ \\
NVNN & $87.38_{\pm10.67}$ & $39.80_{\pm12.83}$ & $45.23_{\pm12.14}$ & $86.94_{\pm8.57}$ \\
NVNV & $83.73_{\pm12.50}$ & $29.55_{\pm13.44}$ & $34.91_{\pm13.78}$ & $89.58_{\pm6.81}$ \\
NVVN & $83.16_{\pm13.11}$ & $29.44_{\pm14.89}$ & $34.70_{\pm14.96}$ & $89.58_{\pm6.81}$ \\
NVVV & $80.35_{\pm17.03}$ & $35.10_{\pm17.84}$ & $42.11_{\pm16.27}$ & $\underline{92.81}_{\pm5.09}$ \\
VNNN & $88.01_{\pm11.46}$ & $39.79_{\pm13.33}$ & $44.92_{\pm12.58}$ & $86.94_{\pm8.57}$ \\
VNNV & $84.77_{\pm13.40}$ & $34.81_{\pm12.86}$ & $40.69_{\pm11.85}$ & $89.58_{\pm6.81}$ \\
VNVN & $84.83_{\pm13.02}$ & $\textbf{46.88}_{\pm14.38}$ & $\underline{54.50}_{\pm11.61}$ & $89.58_{\pm6.81}$ \\
VNVV & $80.36_{\pm17.12}$ & $\underline{42.59}_{\pm18.82}$ & $51.35_{\pm15.51}$ & $\underline{92.81}_{\pm5.09}$ \\
VVNN & $83.54_{\pm13.84}$ & $29.50_{\pm14.50}$ & $34.75_{\pm14.20}$ & $89.58_{\pm6.81}$ \\
VVNV & $79.63_{\pm17.33}$ & $34.03_{\pm17.22}$ & $41.31_{\pm15.98}$ & $\underline{92.81}_{\pm5.09}$ \\
VVVN & $80.38_{\pm18.05}$ & $33.08_{\pm19.02}$ & $39.61_{\pm17.69}$ & $\underline{92.81}_{\pm5.09}$ \\
VVVV & $37.38_{\pm28.54}$ & $24.31_{\pm20.38}$ & $\textbf{63.84}_{\pm27.89}$ & $\textbf{98.17}_{\pm2.48}$ \\

\bottomrule
\end{tabular}
}
\caption{Results of Concepts Coverage and Diverse Rate for each part-of-speech (POS) order of the given four concepts. Higher scores indicate better performance. \textbf{N} denotes nouns, and \textbf{V} denotes verbs. We report the score of mean and standard deviation across 36 LLMs.}
\label{tab:pos_pattern}
\end{table}

We will focus on important aspects in the following sections and defer more discussions to Appendix~\ref{appendix:additional-discussion}.

\subsection{Which Part-of-Speech (POS) Patterns do LLMs Struggle with?}

In GCR tasks, the concepts are categorized into two types: common objects (nouns) and actions (verbs). To investigate the impact of part-of-speech (POS) order on LLM performance, we report the mean and standard deviation for each POS pattern across all 36 LLMs in Table~\ref{tab:pos_pattern}.
Table~\ref{tab:pos_pattern} shows that the NNNN pattern (noun-only) achieves the highest Concepts Coverage. In contrast, the VVVV pattern (verb-only) results in outputs where all concepts are included in only approximately 37\% of cases. Additionally, the VVVV pattern exhibits the highest variance in Concepts Coverage, indicating significant compositional generalization challenges for verb-heavy concept sets in GCR tasks. These challenges also contribute to performance differences among LLMs.
Interestingly, the VVVV pattern achieves the highest scores in both the Ordered Rate and the Diverse Rate. This suggests that, despite difficulties in composing sentences with actions, LLMs actively attempt to do so. Furthermore, when all concepts are included in the output for the VVVV pattern, LLMs demonstrate relatively strong instruction-following capabilities in these scenarios.

\subsection{Impact of Prompt Template Variations on Instruction-Following Performance}

\begin{table}[!t]
\centering
\resizebox{\linewidth}{!}{%
\small
\setlength{\tabcolsep}{2pt}
\rowcolors{2}{gray!10}{white}
\begin{tabular}{@{}llllll@{}}
\toprule
\rowcolor{white}
 & & \multicolumn{3}{c}{Concepts Coverage} & \multicolumn{1}{c}{Diverse} \\ \cmidrule(){3-5}
\rowcolor{white}
\multicolumn{2}{l}{ID} & \multicolumn{1}{c}{w/o order} & \multicolumn{1}{c}{w/ order} & \multicolumn{1}{c}{Ordered Rate} &  \multicolumn{1}{c}{Rate}\\
\midrule
0 & w/o & $81.39_{\pm12.99}$ & $26.76_{\pm 9.30}$ & $33.35_{\pm11.97}$ & $90.77_{\pm 7.16}$ \\
  & w/  & $\textbf{85.05}_{\pm12.57}$ & $\textbf{39.66}_{\pm16.64}$ & $\textbf{45.81}_{\pm15.94}$ & $\textbf{92.13}_{\pm 5.98}$ \\ \midrule
1 & w/o & $83.60_{\pm12.75}$ & $27.67_{\pm 9.27}$ & $33.50_{\pm11.40}$ & $88.86_{\pm 7.79}$ \\
  & w/  & $\textbf{86.34}_{\pm13.06}$ & $\textbf{39.10}_{\pm15.10}$ & $\textbf{44.69}_{\pm14.17}$ & $\textbf{90.90}_{\pm 6.04}$ \\ \midrule
2 & w/o & $77.62_{\pm12.19}$ & $25.52_{\pm10.51}$ & $33.85_{\pm15.31}$ & $\textbf{86.97}_{\pm 9.18}$ \\
  & w/  & $\textbf{83.04}_{\pm12.66}$ & $\textbf{36.91}_{\pm13.67}$ & $\textbf{44.21}_{\pm14.47}$ & $88.62_{\pm 7.37}$ \\ \midrule
3 & w/o & $82.00_{\pm12.25}$ & $22.77_{\pm 8.83}$ & $28.45_{\pm11.88}$ & $89.18_{\pm 8.80}$ \\ 
  & w/  & $\textbf{85.43}_{\pm12.54}$ & $\textbf{31.78}_{\pm15.84}$ & $\textbf{36.62}_{\pm15.63}$ & $\textbf{89.75}_{\pm 8.06}$ \\ \midrule
4 & w/o & $77.09_{\pm12.36}$ & $27.85_{\pm 7.82}$ & $\textbf{37.17}_{\pm12.48}$ & $\textbf{89.79}_{\pm 7.76}$ \\
  & w/  & $\textbf{83.89}_{\pm11.79}$ & $\textbf{30.58}_{\pm 9.97}$ & $36.62_{\pm10.78}$ & $86.77_{\pm 8.97}$ \\ \midrule
5 & w/o & $85.84_{\pm12.65}$ & $17.77_{\pm 5.68}$ & $21.09_{\pm 7.24}$ & $82.02_{\pm13.21}$ \\
  & w/  & $\textbf{87.85}_{\pm12.79}$ & $\textbf{28.88}_{\pm15.98}$ & $\textbf{32.09}_{\pm15.06}$ & $\textbf{84.25}_{\pm11.04}$ \\

\bottomrule
\end{tabular}
}
\caption{The mean and standard deviation of Concepts Coverage and Diverse Rate across 36 LLMs for each prompt. Details of the prompt templates are provided in Appendix~\ref{sec:example-evaluation-templates}. The terms \textbf{w/o} and \textbf{w/} indicate the absence or presence of the phrase \textbf{``in the specified order''} in the prompt template, respectively.}
\label{tab:template-difference}
\end{table}

LLM performance is affected by slight differences in prompt phrasing~\cite{sakai-etal-2024-toward}. To measure the impact of such variations on instruction-following ability, we report the mean and standard deviation of all metrics across 36 LLMs for each of the six prompt templates in Table~\ref{tab:template-difference}.
The results in Table~\ref{tab:template-difference} indicate that while LLM performance varies depending on prompt phrasing, specifying \textbf{``in the specified order''} generally improves w/ order scores across all templates. However, performance differences persist across templates, and some metrics, such as Ordered Rate, do not always improve with this phrasing. 
Furthermore, performance varies significantly among LLMs for each prompt template, making it difficult to identify a universally effective prompt.
These findings highlight the importance of evaluating and reporting averaged results across multiple templates. While prompt-tuning is necessary to achieve optimal performance, the results confirm that LLMs can follow specific instructions when explicitly stated.

\input{1-shot}

\subsection{One-shot Example as Priming}
\label{sec:one-shot}
\paragraph{Motivation.} Humans are known to produce responses influenced by recently encountered information, a phenomenon referred to as priming~\cite{McNamara2005SemanticPP, Bargh2000, zorzi2004computational, 10.1109/TPAMI.2021.3123303, sharma-etal-2024-laying}. For instance, if the preceding input includes the concept set \textit{(apple place tree pick)} and the response begins with \textit{``My favorite words are apple, place, tree, and pick''}, subsequent responses are likely to follow a similar pattern. This tendency is influenced by meta-information, such as word order or response format, rather than the semantic meanings of the words, leading to sentences that start with \textit{``My favorite words are \dots''}.
To simulate this, we evaluated the model’s performance using a one-shot example where the input was a concept set \textit{(apple place tree pick)}, and the generated sentence was \textit{``My favorite words are apple, place, tree, and pick.''}. The evaluation aimed to prime LLMs to produce monotonic outputs by following the template: \textit{``My favorite words are A, B, C, and D''}, where \textit{A}, \textit{B}, \textit{C}, and \textit{D} represent concepts in the order they appear in the set. This priming approach enhances the model’s ability to follow instructions more effectively by leveraging a universal template applicable to any arbitrary set of four concepts. Since this template can generate sentences for any concept set, it serves as a universal \textbf{``ideal example''} shot to improve instruction adherence.

\paragraph{Results.}
Table~\ref{tab:1shot_results} compares the one-shot setting using the ideal example with the zero-shot results of Ordered CommonGen, which assesses the pure inductive ability of LLMs. In Concepts Coverage (w/ order), Table~\ref{tab:1shot_results} highlights a significant priming effect for Llama3.3-70B and Llama3.1-405B, while other models show a considerable drop in scores. In contrast, Concepts Coverage (w/o order) improves for most LLMs, likely due to the influence of the ideal example’s output pattern, which induces reordering.
Focusing on the Diverse Rate, we observe a substantial decrease in scores, while similarity metrics worsen across all models. Additionally, the drop in Distinct scores for most models indicates increased monotonicity in the generated sentences. These findings demonstrate that one-shot examples induce monotonic outputs but reduce the diversity of the generated sentences.
Furthermore, we frequently observed generated sentences deviating from the ideal example patterns and instead producing natural sentences. This suggests that LLMs are not solely primed by the one-shot example but are also influenced by patterns learned from their training data, prioritizing natural text generation.

\paragraph{Summary.}
In conclusion, well-instructed LLMs such as Llama3.3-70B and Llama3.1-405B can simulate priming effects for reasoning. However, most LLMs demonstrate stronger capabilities in generating natural sentences based on patterns learned from their training data. Even for Llama3.1-405B, adherence to all concept sets is not guaranteed, highlighting a persistent challenge. Addressing this issue could enable LLMs to achieve more human-like cognitive text composition, unlocking potential applications such as simulations and beyond.

\section{Conclusion}

In this study, we proposed Ordered CommonGen, a benchmark framework designed to evaluate the instruction-following and compositional generalization capabilities of LLMs. Ordered CommonGen generates permutations of unique concept sets consisting of four concepts and appends the phrase \textbf{``in the specified order''} to the instruction templates to test whether LLMs can compose sentences that include the concepts in the specified order across all permutations. 
The results revealed several key findings: \textit{LLMs understand the intent of instruction prompts; they tend to generate natural sentences while attempting to follow instructions; yet, they often struggle to precisely follow the instructions.}

Furthermore, LLMs produce more diverse outputs when considering concept order but sometimes generate identical sentences even when the concepts are shuffled. Through an analysis of POS patterns and concept sets, we identified specific conditions under which LLMs face challenges in sentence composition. 
Finally, inspired by human perception, we conducted an experiment in Section~\ref{sec:one-shot} using an \textbf{``ideal example''} that demonstrates how a sentence can be composed regardless of the given concepts. The results suggest that well-instructed models have the potential for more human-like, cognitively grounded text generation.

In conclusion, while LLMs demonstrate some degree of compositional generalization capability across arbitrary concept orders, they do not fully exhibit compositionality. In contrast, humans can compose sentences that adhere strictly to specified orders, indicating the potential to create counterexamples for LLMs. Addressing this gap between human and LLM performance remains an important direction for future work and its applications.

\paragraph{Future extensions to other tasks.}
The core idea of our evaluation framework can also be applied to other tasks. For example, it may be extended to poem or story composition based on the order of sentences or scene fragments. Moreover, in the vision domain, e.g., manga or anime, one possible application is composing intermediate images that follow a specified sequence of cells or frames, ensuring the intended development of the story, as well as video tasks. We believe that our evaluation framework is highly versatile and holds great potential for a wide range of creative applications.

\clearpage

\section{Limitations}

\paragraph{Language and Dataset.} %

In this study, we focused on CommonGen~\cite{lin-etal-2020-commongen} and conducted evaluations in English, which limits the generalizability of our findings to other languages, such as Korean~\cite{seo-etal-2022-dog}. To ensure simplicity and consistency in our analysis, we targeted concept sets consisting of four concepts within the CommonGen-lite subset of the CommonGen test data. While increasing the number of concepts could provide more insights~\cite{madaan2023selfrefine}, the associated exponential growth in permutations makes such evaluations highly challenging. Addressing these limitations and exploring tasks with larger concept sets is left for future work.

\paragraph{LLMs.}

Since it is infeasible to evaluate all LLMs, this study focuses on 36 well-known LLMs, a number comparable to other benchmarking studies~\cite{koto-etal-2023-large, koto-etal-2024-arabicmmlu, li-etal-2024-cmmlu, poh-etal-2024-malaymmlu, liu2024reifereevaluatinginstructionfollowingevaluation}. As our research emphasizes instruction-following ability, we conducted evaluations using instruction-tuned LLMs. However, it may be necessary to compare pre-trained base LLMs without instruction-tuning in future studies. Nevertheless, a pilot study with base LLMs showed that these models entirely failed to follow instructions, frequently producing nonsensical sentences, making meaningful evaluation impossible. Therefore, we chose to report results only for models that have undergone at least supervised fine-tuning or further instruction-tuning, ensuring a baseline level of instruction adherence.
Moreover, although variations in tuning~\cite{ismayilzada2025creative} and architecture~\cite{aljaafari2024, aljaafari2025} would ideally be explored, this paper focuses on the evaluation framework and therefore does not include such analyses. Nevertheless, we hope that our framework will contribute to future discussions and improvements in these directions.

\paragraph{Decoding and Prompting.}

Techniques such as prompting~\cite{zhang-etal-2024-improving-diversity, cui-etal-2024-multi}, using external resources~\cite{liu-etal-2023-dimongen, yu-etal-2022-diversifying, hwang-etal-2023-knowledge}, minimum Bayes risk decoding~\cite{deguchi-etal-2024-mbrs, deguchi-etal-2024-centroid, suzgun-etal-2023-follow, jinnai-etal-2024-generating}, or other constrained decoding methods~\cite{sha-2020-gradient, stowe-etal-2022-controlled, willard2023efficientguidedgenerationlarge} might enable the generation of diverse outputs or compositions that adhere to constraints, potentially allowing sentences to be composed from the given concept sets. However, such approaches rely on generation techniques rather than the inductive abilities of the model itself, which is beyond the scope of this study. 
Moreover, we used only the 1-shot setting in Section~\ref{sec:one-shot} to capture the model’s intent, leaving multi-shot settings unexplored.
We focus on the inductive generative commonsense reasoning ability of LLMs, considering their instruction-following capability. This focus aligns with how general-purpose LLMs are typically used by everyday users, ensuring that our evaluation reflects practical usage scenarios.

\paragraph{Evaluation.}

In this study, we did not use traditional reference-based metrics such as BLEU~\cite{papineni-etal-2002-bleu}, ROUGE~\cite{lin-2004-rouge}, METEOR~\cite{banerjee-lavie-2005-meteor},  CIDEr~\cite{conf/cvpr/VedantamZP15}, and SPICE~\cite{10.1007/978-3-319-46454-1_24}, which rely on comparisons with human-annotated gold labels, as employed in original CommonGen paper~\cite{lin-etal-2020-commongen}. Instead, we focused on evaluation methods that do not require labeled data, such as coverage and self-correlation metrics.
We made this choice for several reasons. First, at the time of submission, the test data for CommonGen was no longer publicly available, and we chose to respect the owner’s policy\footnote{\url{https://github.com/INK-USC/CommonGen} and \url{https://hf.co/datasets/allenai/commongen_lite}}. 
Second, comparisons with human-annotated labels in generation tasks have inherent limitations~\cite{reiter-2018-structured, schmidtova-etal-2024-automatic-metrics}. 
Third, human labels for CommonGen are unavailable for all permutations of concept sets, making alignment with our task difficult. Furthermore, the high cost of annotation and scalability challenges led us to avoid reference-based metrics.
Additionally, we did not adopt human evaluation, nor did we use LLM-as-a-judge approaches. These methods are known to have certain issues, are not suited for evaluating large volumes of text, have no established standard method~\cite{wang-etal-2023-chatgpt, gu2024surveyllmasajudge}, and often lead to ambiguous assessment~\cite{clark-etal-2021-thats}.
Considering future research potential, we chose not to employ them. As the primary focus of our study is on instruction-following capability, such evaluations fall outside the scope of this work and are left for future investigation.
Regarding the philosophy behind our framework design, we believe that adopting a low-cost evaluation method is preferable to encourage broader use.
Overall, we believe that our current evaluation framework is sufficient for examining the effectiveness of our approach.

\clearpage
\section*{Ethical Considerations}

We used FLAN templates, released under the Apache License 2.0, and concept sets from CommonGen, provided under the MIT License. Both the templates and the concept sets were modified and extended to suit our research needs. Since these licenses permit such adaptations, our work complies with all licensing requirements. Additionally, our study does not involve potentially harmful content, and all outputs successfully passed the safety moderation filters of the LLMs. Consequently, this research is free from harmful content.

\section*{Acknowledgements}
We thank the anonymous reviewers, area chair, and senior area chair for their valuable comments and suggestions. Their feedback has strengthened this work, encouraged its publication, inspired new research directions, and motivated us.

At the time of preparing this camera-ready version, my (first author) grandmother passed away. I am writing this final revision while on my way to her funeral. Amid the heated academic competition, this moment reminded me of the unwavering support of my family, which had slowly faded from my awareness. I am deeply grateful for their support and I hope to carry her spirit with me as I continue my academic journey. Thank you, and may you rest in eternal peace.

\bibliography{anthology,custom}

\appendix

\section{Detailed LLMs Information}
\label{sec:detailed-llm-information}

\begin{table}[th]
\centering
\resizebox{\linewidth}{!}{%
\footnotesize
\setlength{\tabcolsep}{2pt}
\rowcolors{2}{gray!10}{white}
\begin{tabular}{@{}ll@{}}
\toprule
LLMs &  HuggingFace ID / API Name \\
\midrule
Llama3.2-1B & meta-llama/Llama-3.2-1B-Instruct \\
Llama3.2-3B & meta-llama/Llama-3.2-3B-Instruct \\
Llama3.1-8B & meta-llama/Meta-Llama-3.1-8B-Instruct \\
Llama3.1-70B & meta-llama/Meta-Llama-3.1-70B-Instruct \\
Llama3.3-70B & meta-llama/Llama-3.3-70B-Instruct \\
Llama3.1-405B & meta-llama/Meta-Llama-3.1-405B-Instruct \\ \midrule
Qwen2-0.5B & Qwen/Qwen2-0.5B-Instruct \\
Qwen2-1.5B & Qwen/Qwen2-1.5B-Instruct \\
Qwen2-7B & Qwen/Qwen2-7B-Instruct \\
Qwen2-72B & Qwen/Qwen2-72B-Instruct \\ \midrule
Gemma-2-2B & google/gemma-2-2b-it \\
Gemma-2-9B & google/gemma-2-9b-it \\
Gemma-2-27B & google/gemma-2-27b-it \\ \midrule
Phi-3-mini & microsoft/Phi-3-mini-4k-instruct \\
Phi-3-small & microsoft/Phi-3-small-8k-instruct \\
Phi-3-medium & microsoft/Phi-3-medium-4k-instruct \\ \midrule
Mistral-7B & mistralai/Mistral-7B-Instruct-v0.1 \\
Mistral-small & mistralai/Mistral-Small-Instruct-2409 \\
Mistral-large & mistralai/Mistral-Large-Instruct-2407 \\ \midrule
Mixtral-8x7B & mistralai/Mixtral-8x7B-Instruct-v0.1 \\
Mixtral-8x22B & mistralai/Mixtral-8x22B-Instruct-v0.1 \\ \midrule
Qwen2.5-0.5B & Qwen/Qwen2.5-0.5B-Instruct \\
Qwen2.5-1.5B & Qwen/Qwen2.5-1.5B-Instruct \\
Qwen2.5-3B & Qwen/Qwen2.5-3B-Instruct \\
Qwen2.5-7B & Qwen/Qwen2.5-7B-Instruct \\
Qwen2.5-14B & Qwen/Qwen2.5-14B-Instruct \\
Qwen2.5-32B & Qwen/Qwen2.5-32B-Instruct \\
Qwen2.5-72B & Qwen/Qwen2.5-72B-Instruct \\ \midrule
T\"ulu3-8B & allenai/Llama-3.1-Tulu-3-8B \\
T\"ulu3-70B & allenai/Llama-3.1-Tulu-3-70B \\ \midrule
OMLo2-7B & allenai/OLMo-2-1124-7B \\
OMLo2-13B & allenai/OLMo-2-1124-13B \\ \midrule
GPT-3.5 & OpenAI/gpt-3.5-turbo-0125 \\
GPT-4o & OpenAI/gpt-4o-2024-05-13 \\ \midrule
Gemini-Flash & Gemini/gemini-1.5-flash-001 \\
Gemini-Pro & Gemini/gemini-1.5-pro-001 \\
\bottomrule
\end{tabular}
}
\caption{Details of the LLMs for the experiments.}
\label{tab:models}
\end{table}

Table~\ref{tab:models} shows the source information of each LLM.
We used the Transformers~\cite{wolf-etal-2020-transformers} and bitsandbytes~\cite{dettmers2022llmint8} libraries for the inference of open LLMs. 
We used a single A6000 48GB GPU for most open LLMs, while Mistral-large was run on two A6000 48GB GPUs, and Llama3.1-405B was run on eight A6000 48GB GPUs.
Note that we used only instruction-tuned LLMs because base models do not follow instructions precisely, and this aligns with our research purpose, which is to evaluate instruction-following capabilities.
For instruction-tuned LLMs, we conducted inference by applying their specific chat templates. As a system prompt, we included the phrase: \textit{''You are a helpful assistant. Please generate only an answer.''}. Upon randomly sampling 50 outputs per model and manually inspecting them, we observed no notable generation of supplementary information or flavor text. Therefore, the generated text was directly used for evaluation. The same procedure was applied to proprietary models.

\section{Examples of Instruction Template}
\label{sec:example-evaluation-templates}

\begin{table}[t]
\centering
\small
\setlength{\tabcolsep}{3.5pt}
\rowcolors{2}{gray!10}{white}
\resizebox{\linewidth}{!}{%
\begin{tabular}{ll}
\toprule
ID& Instruction Template \\
\midrule

0 & \makecell[{{p{7cm}}}t]{
Concepts: \{concepts\} 
\\ \ \\
Write a sentence that includes all these words \textbf{in the specified order}.
\\ \ \\
Answer:
} \\ \midrule

1 & \makecell[{{p{7cm}}}t]{
Keywords: \{concepts\}
\\ \ \\
What is a sentence that includes all these keywords \textbf{in the specified order}?
\\ \ \\
Answer:
} \\ \midrule

2 & \makecell[{{p{7cm}}}t]{
Here are some concepts: \{concepts\}
\\ \ \\
What is a sentence about these concepts \textbf{in the specified order}?
\\ \ \\
Answer:
} \\ \midrule

3 & \makecell[{{p{7cm}}}t]{
Produce a sentence which mentions all of these concepts \textbf{in the specified order}: \{concepts\}
\\ \ \\
Answer:
} \\ \midrule

4 & \makecell[{{p{7cm}}}t]{
Write a sentence about the following things \textbf{in the specified order}:
\\ \ \\
\{concepts\}
\\ \ \\
Answer:
} \\ \midrule

5 & \makecell[{{p{7cm}}}t]{
Generate a sentence that includes all the following words \textbf{in the specified order}: \{concepts\}
\\ \ \\
Answer:
} \\

\bottomrule
\end{tabular}
}
\caption{The instruction templates for our experiments. We created it based on the FLAN template and inserted it \textbf{``in the specified order''} in the bold one. Each concept from the concepts set is assigned to the curly brackets as space-separated values.}
\label{tab:template-details}
\end{table}

Table~\ref{tab:template-details} shows the instruction template used in the Ordered CommonGen experiment framework. Each concept from the concepts set is assigned to the curly brackets \textbf{\{concepts\}} as space-separated values. 
Following the findings of \citet{zhang-etal-2023-learning-predict}, which demonstrated slightly better performance with space-separated values and were corroborated by our pilot study, we opted for space separation. This separation method is also reasonable from a mechanistic perspective: when considering the ID token sequence converted by tokenization of LLMs, separating concepts with commas or other delimiters introduces additional tokens between concepts, which could affect performance.

\section{Additional Discussions}
\label{appendix:additional-discussion}

\subsection{Performance of Reasoning Models}

\begin{table}[!t]
\centering
\small
\setlength{\tabcolsep}{4pt}
\rowcolors{2}{gray!10}{white}
\begin{tabular}{lllll}
\toprule
\rowcolor{white}
& \multicolumn{1}{c}{w/o order} & \multicolumn{1}{c}{w/ order} & \multicolumn{1}{c}{Ordered Rate}\\
\midrule
Qwen2-7B & $\textbf{89.28}_{\text{(+5.16)}}$ & $\textbf{34.22}_{\text{(+7.96)}}$ & $38.33_{\text{(+7.11)}}$ \\
Macro-o1 & $79.88_{\text{(+6.88})}$ & $33.27_{\text{(+6.90)}}$ & $\textbf{41.65}_{\text{(+5.53)}}$ \\ \midrule
GPT-4o-mini & $87.72_{\text{(+4.99)}}$ & $43.27_{\text{(+16.58})}$ & $49.33_{\text{(+17.07)}}$ \\
o1-mini & $\textbf{99.09}_{\text{(+2.80)}}$ & $\textbf{84.70}_{\text{(+70.38)}}$ & $\textbf{85.48}_{\text{(+70.60)}}$ \\
\bottomrule
\end{tabular}

\caption{Comparison of evaluation results for the generated sentences between each base model and reasoning model. We use the single template with ID 0 from Table~\ref{tab:template-details} in Appendix~\ref{sec:example-evaluation-templates}. We report the Concepts Coverage metrics, where higher scores indicate better performance, with performance differences from the original CommonGen, which does not consider concept order, shown in parentheses. The upper row of each section represents the base model, while the lower row represents its corresponding reasoning model.}
\label{tab:reasoning-model}
\end{table}

Reasoning using Chain of Thought (CoT)~\cite{wei2022chain, chu-etal-2024-navigate} prompts enhances the reasoning capabilities of LLMs. Recently, reasoning models~\cite{deepseek1, openai2024o1mini} have emerged that achieve comparable capabilities through training rather than prompt-based control. Due to variations in CoT reasoning, unified evaluation is challenging; therefore, we evaluated using reasoning models. Table~\ref{tab:reasoning-model} compares the evaluation results for Qwen2-7B~\cite{yang2024qwen2technicalreport}\footnote{\url{https://hf.co/Qwen/Qwen2-7B-Instruct}} and its reasoning model Marco-o1~\cite{zhao2024marcoo1openreasoningmodels}\footnote{\url{https://hf.co/AIDC-AI/Marco-o1}}, as well as GPT-4o-mini~\cite{openai2024gpt4technicalreport}\footnote{gpt-4o-mini-2024-07-18} and its reasoning model o1-mini~\cite{openai2024o1mini}\footnote{o1-mini-2024-09-12}\footnote{We attempted experiments with o1 (o1-2024-09-12) but could not complete them due to financial cost constraints. However, we observed instances where the model generated outputs that did not follow the specified order. This suggests that the same limitations apply to o1, indicating that it does not yet guarantee 100\% adherence to instructions.}. As shown in Table~\ref{tab:reasoning-model}, reasoning models, particularly o1-mini, demonstrate significant improvements in Ordered Rate compared to their base models. While Marco-o1 shows a decrease in scores like w/ order, its Ordered Rate improves, reflecting enhanced inference capabilities through effective reasoning. These results indicate that reasoning is effective; however, even these models do not guarantee 100\% adherence. Thus, while reasoning enhances inductive generative commonsense reasoning in LLMs, a noticeable gap remains between these models and human capabilities in composing sentences with arbitrary orders.

\begin{figure*}[!t]
\centering
  \includegraphics[width=.975\linewidth]{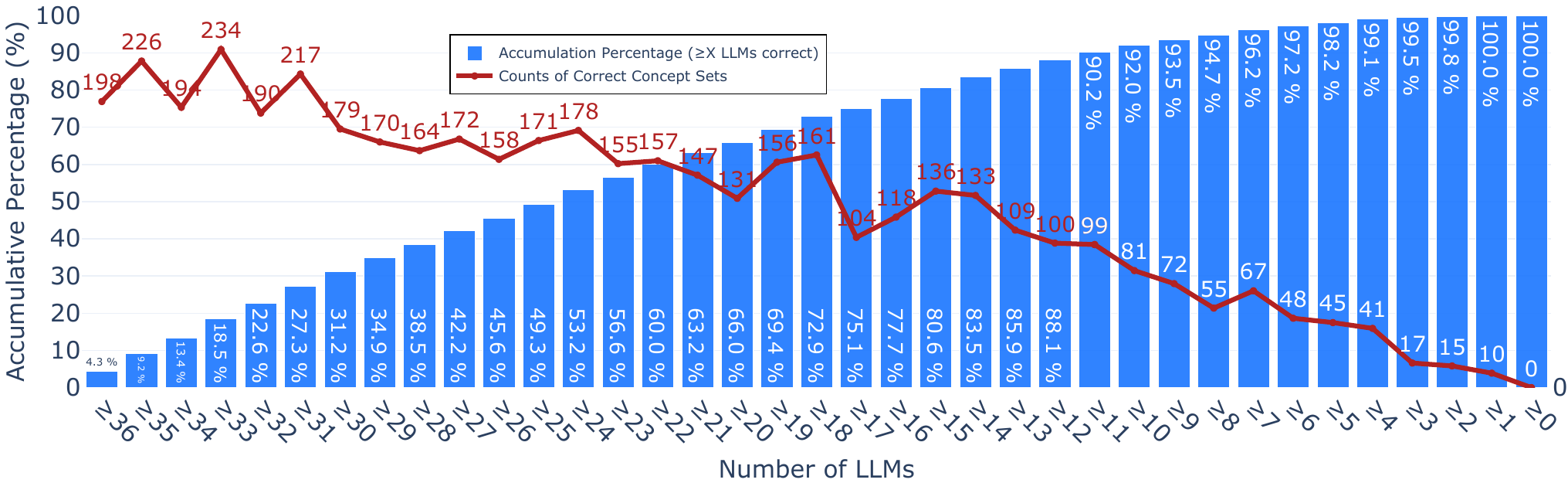}
  \caption{The distribution of all 4,608 concept sets successfully composed by all 36 LLMs using any of the six templates. The x-axis represents the number of LLMs that successfully composed given concept sets, while the y-axis shows the accumulative percentage. The line graph indicates the number of concept sets in each bin.}
  \label{fig:can_compose}
\end{figure*}

\begin{figure*}[!t]
\centering
  \includegraphics[width=.975\linewidth]{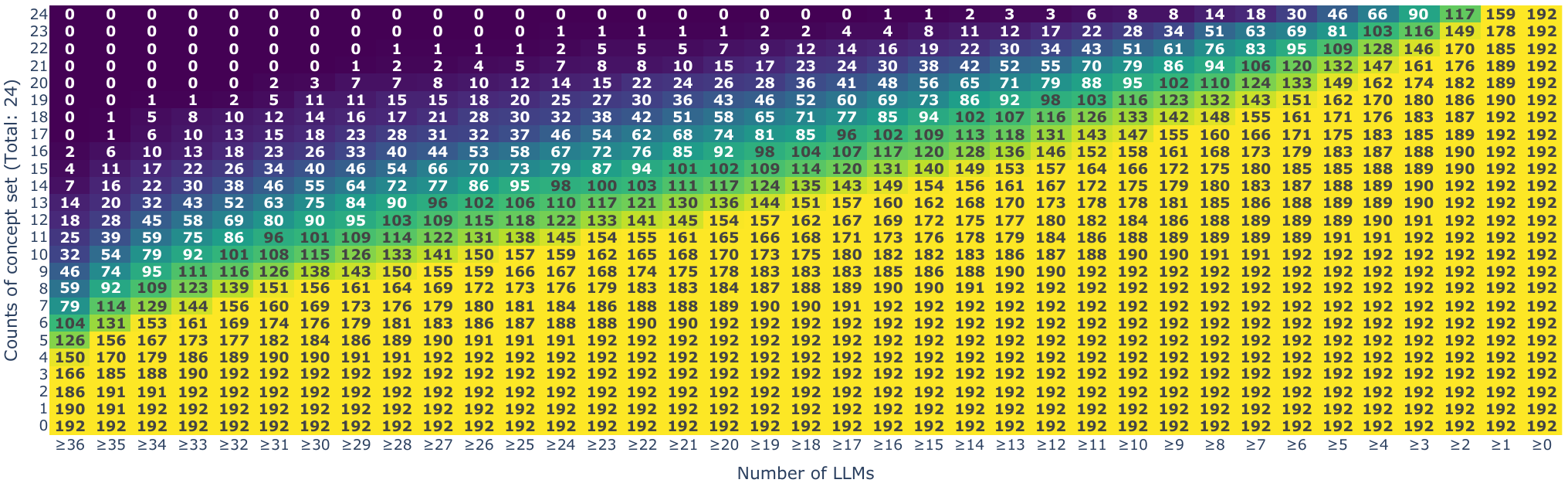}
  \caption{The distribution of successfully composed permutations of concept sets by all 36 LLMs using any of the six templates. The x-axis represents the number of LLMs that successfully composed each permuted concept set, while the y-axis shows the counts of concept sets. A total of 192 unique concept sets is evaluated, with a maximum of 24 permutations per set ($4! = 24$). Black numbers represent the majority of concept sets, while white numbers indicate the minority.}
  \label{fig:matrix}
\end{figure*}

\subsection{Future Evaluation Direction}
In this evaluation framework, as the primary focus of our study is on instruction-following capability, we prioritized self-evaluation-based methods. This allowed us to effectively assess model performance even in open-ended generation tasks. However, if the goal shifts toward generating more semantically and syntactically natural sentences, additional evaluation methods may be needed. For example, it will be important to estimate grammatical quality by detecting grammatical errors~\cite{sakai2025imparaged, goto2025gecmetrics, goto2025rethink, maeda-etal-2022-impara, yoshimura-etal-2020-reference}, or to assess linguistic acceptabilities~\cite{warstadt-etal-2019-neural, tjuatja-etal-2025-goes, ide-etal-2025-make}.
We share these as future directions and challenges toward advancing reference-free evaluation of text generation, as well as potential extensions of our Ordered CommonGen task.

\subsection{Which Concept Sets Are Difficult for LLMs to Compose?}
\label{sec:case-study}

Figure~\ref{fig:can_compose} shows the distribution of the number and proportion of LLMs that successfully composed sentences from each of the 4,608 concept sets in Ordered CommonGen. As shown in Figure~\ref{fig:can_compose}, every concept set was composed by at least one LLM. 
These results reveal that while at least one LLM can compose sentences using the given concepts in the specified order, compositional generalization abilities vary significantly across models, indicating instability in this capability.
Figure~\ref{fig:matrix} shows which permutations of concept sets are the most difficult for LLMs. The results indicate that at least one LLM was able to compose sentences for all permutations of 159 concept sets. However, for 33 concept sets, no LLM could successfully compose sentences for any of their permutations. Notably, if a success rate of 75\% (18 out of 24 permutations) is acceptable, at least one LLM could successfully compose sentences for all concept sets. These findings suggest that while LLMs can compose sentences for a wide range of concept order permutations, they do not consistently adhere to instructions across all permutations, further highlighting the overall instability and limitations in compositional generalization.

\end{document}

%% file: main_result.tex
\begin{table*}[!t]
\centering
\resizebox{\linewidth}{!}{%
\small
\setlength{\tabcolsep}{2pt}
\rowcolors{2}{gray!10}{white}
\begin{tabular}{@{}lllllllll@{}}
\toprule
\rowcolor{white}
 & \multicolumn{3}{c}{Concepts Coverage ($\uparrow$)} & \multicolumn{2}{c}{Similarlity ($\downarrow$)} & \multicolumn{2}{c}{Diversity ($\uparrow$)} & \multicolumn{1}{@{}c@{}}{\multirow{2}{*}{Perplexity ($\downarrow$)}} \\ 
 \cmidrule(lr){2-4} \cmidrule(l){5-6} \cmidrule(lr){7-8} %
 \rowcolor{white}
 & \multicolumn{1}{c}{w/o order} & \multicolumn{1}{c}{w/ order} & \multicolumn{1}{c}{Ordered Rate} & \multicolumn{1}{c}{pBLEU} & \multicolumn{1}{c}{pBLEURT} & \multicolumn{1}{c}{Distinct} & \multicolumn{1}{c}{Diverse Rate} & \\ \midrule

Llama3.2-1B & $61.82_{\text{(+2.56)}}$ & $17.22_{\text{( -1.02)}}$ & $27.86_{\text{( -2.92)}}$ & $17.62_{\text{(+2.37)}}$ & $41.89_{\text{(+1.95)}}$ & $92.79_{\text{( -1.17)}}$ & $93.66_{\text{( -2.35)}}$ & $52.46_{\text{(+7.95)}}$ \\
Llama3.2-3B & $63.81_{\text{(+2.64)}}$ & $19.47_{\text{(+1.33)}}$ & $30.52_{\text{(+0.86)}}$ & $18.49_{\text{(+2.02)}}$ & $44.38_{\text{(+3.08)}}$ & $94.18_{\text{( -0.84)}}$ & $90.53_{\text{( -2.88)}}$ & $53.85_{\text{(+8.47)}}$ \\
Llama3.1-8B & $72.10_{\text{(+3.32)}}$ & $23.35_{\text{(+3.80)}}$ & $32.39_{\text{(+3.96)}}$ & $19.61_{\text{(+0.66)}}$ & $47.30_{\text{(+2.83)}}$ & $93.45_{\text{( -1.14)}}$ & $87.91_{\text{( -2.24)}}$ & $61.72_{\text{(+13.47)}}$ \\
Llama3.1-70B & $95.69_{\text{(+4.12)}}$ & $41.19_{\text{(+25.66)}}$ & $43.04_{\text{(+26.08)}}$ & $21.89_{\text{( -7.30)}}$ & $52.40_{\text{( -1.92)}}$ & $93.03_{\text{( -0.61)}}$ & $83.84_{\text{(+8.73)}}$ & $60.24_{\text{(+11.39)}}$ \\
Llama3.3-70B & $\underline{97.25}_{\text{(+2.17)}}$ & $\underline{66.79}_{\text{(\underline{+47.34})}}$ & $\underline{68.68}_{\text{(\underline{+48.22})}}$ & $15.24_{\text{(\textbf{ -11.17})}}$ & $48.04_{\text{( -4.45)}}$ & $93.53_{\text{( -0.50)}}$ & $94.70_{\text{(\underline{+14.13})}}$ & $67.50_{\text{(+21.19)}}$ \\
Llama3.1-405B & $\textbf{98.91}_{\text{(+2.84)}}$ & $\textbf{74.44}_{\text{(\textbf{+55.41})}}$ & $\textbf{75.26}_{\text{(\textbf{+55.46})}}$ & $12.54_{\text{(\underline{ -11.12})}}$ & $42.90_{\text{(\textbf{ -5.72})}}$ & $94.81_{\text{(\underline{ -0.20})}}$ & $\underline{98.28}_{\text{(+13.43)}}$ & $61.49_{\text{(+17.73)}}$ \\ \midrule
Qwen2-0.5B & $53.78_{\text{( -0.61)}}$ & $30.84_{\text{( -0.89)}}$ & $57.34_{\text{( -0.98)}}$ & $\textbf{10.48}_{\text{(+0.57)}}$ & $\underline{39.61}_{\text{(+1.35)}}$ & $93.31_{\text{( -0.72)}}$ & $96.60_{\text{( -1.01)}}$ & $171.27_{\text{(+81.35)}}$ \\
Qwen2-1.5B & $73.06_{\text{(+0.91)}}$ & $24.12_{\text{( -0.65)}}$ & $33.01_{\text{( -1.32)}}$ & $15.72_{\text{(+1.27)}}$ & $43.70_{\text{(+1.17)}}$ & $93.71_{\text{( -0.32)}}$ & $91.81_{\text{( -2.86)}}$ & $53.52_{\text{(\underline{+2.06})}}$ \\
Qwen2-7B & $88.52_{\text{(+4.16)}}$ & $27.49_{\text{(+2.72)}}$ & $31.06_{\text{(+1.70)}}$ & $13.72_{\text{( -0.08)}}$ & $43.93_{\text{(+0.98)}}$ & $94.26_{\text{( -0.69)}}$ & $95.49_{\text{( -0.82)}}$ & $70.46_{\text{(+12.74)}}$ \\
Qwen2-72B & $94.99_{\text{(+3.51)}}$ & $32.09_{\text{(+7.13)}}$ & $33.78_{\text{(+6.50)}}$ & $19.13_{\text{( -1.53)}}$ & $48.58_{\text{(+0.25)}}$ & $93.47_{\text{( -0.72)}}$ & $85.49_{\text{(+0.35)}}$ & $\textbf{48.68}_{\text{(+3.45)}}$ \\ \midrule
Gemma2-2B & $80.07_{\text{(\textbf{+10.79})}}$ & $23.66_{\text{(+0.44)}}$ & $29.55_{\text{( -3.97)}}$ & $21.28_{\text{(+3.84)}}$ & $51.04_{\text{(+5.22)}}$ & $91.09_{\text{( -1.69)}}$ & $82.45_{\text{( -7.27)}}$ & $159.99_{\text{(+80.40)}}$ \\
Gemma2-9B & $90.10_{\text{(+6.76)}}$ & $23.56_{\text{(+3.82)}}$ & $26.15_{\text{(+2.47)}}$ & $23.95_{\text{(+1.06)}}$ & $53.74_{\text{(+2.53)}}$ & $91.87_{\text{( -0.63)}}$ & $78.94_{\text{( -1.86)}}$ & $88.74_{\text{(+11.56)}}$ \\
Gemma2-27B & $88.49_{\text{(\underline{+8.58})}}$ & $28.24_{\text{(+7.73)}}$ & $31.91_{\text{(+6.24)}}$ & $22.14_{\text{( -0.86)}}$ & $51.35_{\text{(+0.88)}}$ & $92.16_{\text{( -0.58)}}$ & $80.84_{\text{(+0.42)}}$ & $94.27_{\text{(+17.05)}}$ \\ \midrule
Phi3-mini & $79.85_{\text{(+4.90)}}$ & $49.54_{\text{(+2.73)}}$ & $62.04_{\text{( -0.41)}}$ & $\underline{10.72}_{\text{(+0.04)}}$ & $41.70_{\text{(+0.39)}}$ & $94.36_{\text{( -0.56)}}$ & $97.53_{\text{( -0.72)}}$ & $100.71_{\text{(+11.19)}}$ \\
Phi3-small & $89.79_{\text{(+3.38)}}$ & $49.93_{\text{(+14.13)}}$ & $55.61_{\text{(+14.17)}}$ & $10.99_{\text{( -1.22)}}$ & $40.81_{\text{( -0.28)}}$ & $94.22_{\text{( -0.63)}}$ & $97.01_{\text{( -0.38)}}$ & $77.02_{\text{(+15.27)}}$ \\
Phi3-medium & $85.84_{\text{(+1.77)}}$ & $50.21_{\text{(+6.85)}}$ & $58.49_{\text{(+6.92)}}$ & $10.91_{\text{( -0.77)}}$ & $39.69_{\text{( -0.46)}}$ & $\textbf{95.16}_{\text{(\textbf{ -0.07})}}$ & $98.23_{\text{(+0.48)}}$ & $80.62_{\text{(+4.38)}}$ \\ \midrule
Mistral-7B & $81.26_{\text{(+3.82)}}$ & $40.71_{\text{( -2.42)}}$ & $50.10_{\text{( -5.60)}}$ & $18.13_{\text{(+2.14)}}$ & $48.79_{\text{(+2.66)}}$ & $92.56_{\text{( -1.36)}}$ & $86.49_{\text{( -5.64)}}$ & $238.96_{\text{(+157.60)}}$ \\
Mistral-small & $95.36_{\text{(+2.68)}}$ & $37.12_{\text{(+17.12)}}$ & $38.92_{\text{(+17.34)}}$ & $22.93_{\text{( -5.86)}}$ & $55.55_{\text{( -2.86)}}$ & $91.64_{\text{( -0.35)}}$ & $78.39_{\text{(+7.90)}}$ & $97.88_{\text{(+21.38)}}$ \\
Mistral-large & $87.07_{\text{(+5.05)}}$ & $23.67_{\text{( -0.73)}}$ & $27.19_{\text{( -2.56)}}$ & $25.27_{\text{(+2.80)}}$ & $54.61_{\text{(+3.13)}}$ & $92.58_{\text{( -0.66)}}$ & $76.54_{\text{( -4.98)}}$ & $94.91_{\text{(+9.18)}}$ \\ \midrule
Mixtral-8x7B & $77.36_{\text{(+2.90)}}$ & $19.67_{\text{(+0.09)}}$ & $25.43_{\text{( -0.87)}}$ & $11.29_{\text{(+0.36)}}$ & $\textbf{38.50}_{\text{(+0.31)}}$ & $\underline{95.06}_{\text{( -0.31)}}$ & $\textbf{98.82}_{\text{( -0.17)}}$ & $\underline{52.35}_{\text{(\textbf{+1.51})}}$ \\
Mixtral-8x22B & $90.96_{\text{(+4.52)}}$ & $36.18_{\text{(+9.11)}}$ & $39.77_{\text{(+8.46)}}$ & $13.40_{\text{( -0.03)}}$ & $42.34_{\text{(+0.68)}}$ & $94.55_{\text{( -0.43)}}$ & $96.08_{\text{( -0.58)}}$ & $56.51_{\text{(+2.19)}}$ \\ \midrule
Qwen2.5-0.5B & $64.50_{\text{(+3.27)}}$ & $28.12_{\text{(+0.86)}}$ & $43.60_{\text{( -0.92)}}$ & $10.86_{\text{( -0.09)}}$ & $40.38_{\text{( -0.59)}}$ & $93.27_{\text{(\textbf{ -0.07})}}$ & $97.12_{\text{(+0.15)}}$ & $94.22_{\text{(+19.41)}}$ \\
Qwen2.5-1.5B & $57.09_{\text{(+5.97)}}$ & $14.67_{\text{(+0.02)}}$ & $25.69_{\text{( -2.96)}}$ & $14.65_{\text{(+0.80)}}$ & $43.22_{\text{(+2.39)}}$ & $94.28_{\text{( -0.63)}}$ & $91.71_{\text{( -1.64)}}$ & $64.83_{\text{(+7.50)}}$ \\
Qwen2.5-3B & $86.01_{\text{(+4.98)}}$ & $22.60_{\text{(+3.36)}}$ & $26.28_{\text{(+2.53)}}$ & $14.85_{\text{(+1.22)}}$ & $47.51_{\text{(+2.38)}}$ & $91.94_{\text{( -0.99)}}$ & $92.75_{\text{( -2.79)}}$ & $115.35_{\text{(+26.75)}}$ \\
Qwen2.5-7B & $87.80_{\text{(+4.46)}}$ & $28.58_{\text{(+6.26)}}$ & $32.55_{\text{(+5.76)}}$ & $15.19_{\text{(+0.59)}}$ & $45.97_{\text{(+1.76)}}$ & $92.94_{\text{( -0.73)}}$ & $92.35_{\text{( -1.56)}}$ & $75.97_{\text{(+6.24)}}$ \\
Qwen2.5-14B & $94.49_{\text{(+3.84)}}$ & $40.65_{\text{(+15.76)}}$ & $43.02_{\text{(+15.56)}}$ & $13.96_{\text{( -2.15)}}$ & $44.38_{\text{( -0.14)}}$ & $93.10_{\text{( -0.53)}}$ & $93.21_{\text{(+1.31)}}$ & $68.24_{\text{(+8.90)}}$ \\
Qwen2.5-32B & $96.50_{\text{(+3.18)}}$ & $47.23_{\text{(+22.82)}}$ & $48.94_{\text{(+22.78)}}$ & $12.94_{\text{( -4.11)}}$ & $43.56_{\text{( -2.85)}}$ & $92.80_{\text{( -0.51)}}$ & $93.20_{\text{(+3.18)}}$ & $71.03_{\text{(+7.20)}}$ \\
Qwen2.5-72B & $97.17_{\text{(+3.91)}}$ & $50.26_{\text{(+24.66)}}$ & $51.73_{\text{(+24.27)}}$ & $17.24_{\text{( -5.01)}}$ & $47.98_{\text{( -1.64)}}$ & $93.19_{\text{( -0.50)}}$ & $88.51_{\text{(+6.41)}}$ & $72.02_{\text{(+13.93)}}$ \\ \midrule
OLMo2-7B & $91.11_{\text{(+2.09)}}$ & $28.10_{\text{( -3.97)}}$ & $30.84_{\text{( -5.18)}}$ & $19.05_{\text{(+2.21)}}$ & $52.44_{\text{(+2.82)}}$ & $91.65_{\text{( -0.69)}}$ & $85.25_{\text{( -3.67)}}$ & $112.93_{\text{(+15.01)}}$ \\
OLMo2-13B & $95.22_{\text{(+3.29)}}$ & $27.20_{\text{(+1.68)}}$ & $28.56_{\text{(+0.80)}}$ & $24.33_{\text{(+1.51)}}$ & $58.05_{\text{(+2.69)}}$ & $90.01_{\text{( -1.32)}}$ & $77.20_{\text{( -3.28)}}$ & $433.50_{\text{(+301.35)}}$ \\ \midrule
T\"ulu3-8B & $89.79_{\text{(+4.04)}}$ & $27.55_{\text{( -1.63)}}$ & $30.69_{\text{( -3.35)}}$ & $16.09_{\text{(+1.99)}}$ & $45.97_{\text{(+3.48)}}$ & $92.61_{\text{( -1.16)}}$ & $90.42_{\text{( -3.39)}}$ & $103.14_{\text{(+34.83)}}$ \\
T\"ulu3-70B & $93.46_{\text{(+3.58)}}$ & $33.46_{\text{(+12.28)}}$ & $35.80_{\text{(+12.24)}}$ & $21.55_{\text{( -3.26)}}$ & $52.25_{\text{( -0.44)}}$ & $92.23_{\text{( -0.57)}}$ & $80.04_{\text{(+3.52)}}$ & $79.78_{\text{(+22.72)}}$ \\ \midrule
GPT-3.5 & $92.74_{\text{(+2.15)}}$ & $34.14_{\text{(+8.22)}}$ & $36.82_{\text{(+8.20)}}$ & $23.56_{\text{( -1.75)}}$ & $56.83_{\text{(+0.34)}}$ & $90.88_{\text{( -0.99)}}$ & $76.16_{\text{(+0.68)}}$ & $236.08_{\text{(+155.67)}}$ \\
GPT-4o & $96.70_{\text{(+3.12)}}$ & $53.34_{\text{(+30.25)}}$ & $55.16_{\text{(+30.49)}}$ & $19.95_{\text{( -9.37)}}$ & $50.94_{\text{(\underline{ -4.80})}}$ & $92.91_{\text{( -0.25)}}$ & $86.51_{\text{(\textbf{+14.47})}}$ & $58.15_{\text{(+7.27)}}$ \\ \midrule
Gemini-Flash & $91.79_{\text{(+7.68)}}$ & $33.29_{\text{(+14.51)}}$ & $36.27_{\text{(+13.94)}}$ & $24.03_{\text{( -1.78)}}$ & $56.16_{\text{(+2.08)}}$ & $91.37_{\text{( -0.89)}}$ & $75.92_{\text{(+1.40)}}$ & $92.75_{\text{(+29.04)}}$ \\
Gemini-Pro & $89.16_{\text{(+8.03)}}$ & $32.68_{\text{(+16.52)}}$ & $36.66_{\text{(+16.74)}}$ & $18.67_{\text{( -2.79)}}$ & $50.88_{\text{( -0.11)}}$ & $91.51_{\text{( -0.82)}}$ & $84.48_{\text{(+2.40)}}$ & $117.09_{\text{(+41.66)}}$ \\
\bottomrule
\end{tabular}
}
\caption{Evaluation results for the generated sentences of each LLM. The average scores for Ordered CommonGen, where the order of concepts is specified, are reported across all six templates as the main results, with performance differences from the original CommonGen, which does not consider concept order, shown in parentheses. The bold scores highlight the LLM with the best performance for each metric, while the underlined scores indicate the second-best. Higher scores indicate better performance for Concepts Coverage and Corpus-level Diversity, whereas lower scores indicate better performance for Sentence-level Similarity and Perplexity.}
\label{tab:main_results}
\vspace{-0.01em}
\end{table*}

%% file: 1-shot.tex
\begin{table*}[!t]
\centering
\resizebox{\linewidth}{!}{%
\small
\setlength{\tabcolsep}{2pt}
\rowcolors{2}{gray!10}{white}
\begin{tabular}{@{}lllllllll@{}}
\toprule
\rowcolor{white}
 & \multicolumn{3}{c}{Concepts Coverage ($\uparrow$)} & \multicolumn{2}{c}{Similarlity ($\downarrow$)} & \multicolumn{2}{c}{Diversity ($\uparrow$)} & \multicolumn{1}{@{}c@{}}{\multirow{2}{*}{Perplexity ($\downarrow$)}} \\ 
 \cmidrule(lr){2-4} \cmidrule(l){5-6} \cmidrule(lr){7-8} %
 \rowcolor{white}
 & \multicolumn{1}{c}{w/o order} & \multicolumn{1}{c}{w/ order} & \multicolumn{1}{c}{Ordered Rate} & \multicolumn{1}{c}{pBLEU} & \multicolumn{1}{c}{pBLEURT} & \multicolumn{1}{c}{Distinct} & \multicolumn{1}{c}{Diverse Rate} & \\ \midrule

Llama3.2-1B & $67.05_{\text{(+5.23)}}$ & $13.82_{\text{( -3.40)}}$ & $20.61_{\text{( -7.25)}}$ & $23.60_{\text{(+5.98)}}$ & $47.63_{\text{(+5.73)}}$ & $92.75_{\text{( -0.04)}}$ & $85.52_{\text{( -8.14)}}$ & $\textbf{43.97}_{\text{( -8.49)}}$ \\
Llama3.2-3B & $69.67_{\text{(+5.86)}}$ & $13.57_{\text{( -5.90)}}$ & $19.48_{\text{( -11.03)}}$ & $28.22_{\text{(+9.73)}}$ & $52.50_{\text{(+8.12)}}$ & $93.84_{\text{( -0.34)}}$ & $78.46_{\text{( -12.07)}}$ & $48.03_{\text{( -5.82)}}$ \\
Llama3.1-8B & $85.26_{\text{(\textbf{+13.16})}}$ & $18.31_{\text{( -5.05)}}$ & $21.47_{\text{( -10.92)}}$ & $28.36_{\text{(+8.76)}}$ & $54.16_{\text{(+6.86)}}$ & $93.29_{\text{( -0.15)}}$ & $76.43_{\text{( -11.48)}}$ & $48.22_{\text{( -13.50)}}$ \\
Llama3.1-70B & $98.57_{\text{(+2.88)}}$ & $41.62_{\text{(+0.43)}}$ & $42.22_{\text{( -0.82)}}$ & $27.06_{\text{(+5.17)}}$ & $56.12_{\text{(+3.72)}}$ & $93.00_{\text{( -0.03)}}$ & $78.43_{\text{( -5.41)}}$ & $49.63_{\text{( -10.61)}}$ \\
Llama3.3-70B & $\underline{99.01}_{\text{(+1.76)}}$ & $\underline{75.81}_{\text{(\underline{+9.03})}}$ & $\underline{76.58}_{\text{(\underline{+7.90})}}$ & $18.01_{\text{(+2.77)}}$ & $49.72_{\text{(+1.68)}}$ & $93.71_{\text{(+0.19)}}$ & $\underline{95.72}_{\text{(\underline{+1.01})}}$ & $55.80_{\text{( -11.70)}}$ \\
Llama3.1-405B & $\textbf{99.56}_{\text{(+0.65)}}$ & $\textbf{87.79}_{\text{(\textbf{+13.35})}}$ & $\textbf{88.18}_{\text{(\textbf{+12.92})}}$ & $14.65_{\text{(+2.12)}}$ & $46.01_{\text{(+3.11)}}$ & $\underline{94.10}_{\text{( -0.71)}}$ & $\textbf{99.37}_{\text{(\textbf{+1.09})}}$ & $65.66_{\text{(+4.18)}}$ \\ \midrule
Qwen2-0.5B & $35.39_{\text{( -18.40)}}$ & $16.32_{\text{( -14.52)}}$ & $46.12_{\text{( -11.23)}}$ & $15.16_{\text{(+4.68)}}$ & $43.73_{\text{(+4.12)}}$ & $93.10_{\text{( -0.20)}}$ & $91.31_{\text{( -5.30)}}$ & $57.46_{\text{( -113.81)}}$ \\
Qwen2-1.5B & $63.50_{\text{( -9.57)}}$ & $13.06_{\text{( -11.05)}}$ & $20.57_{\text{( -12.43)}}$ & $21.87_{\text{(+6.16)}}$ & $50.41_{\text{(+6.71)}}$ & $92.99_{\text{( -0.72)}}$ & $84.31_{\text{( -7.50)}}$ & $49.56_{\text{( -3.96)}}$ \\
Qwen2-7B & $89.82_{\text{(+1.30)}}$ & $17.44_{\text{( -10.05)}}$ & $19.41_{\text{( -11.64)}}$ & $19.75_{\text{(+6.04)}}$ & $48.71_{\text{(+4.79)}}$ & $94.00_{\text{( -0.25)}}$ & $89.68_{\text{( -5.82)}}$ & $52.69_{\text{( -17.76)}}$ \\
Qwen2-72B & $96.77_{\text{(+1.79)}}$ & $27.97_{\text{( -4.12)}}$ & $28.90_{\text{( -4.88)}}$ & $24.28_{\text{(+5.16)}}$ & $53.95_{\text{(+5.37)}}$ & $92.80_{\text{( -0.67)}}$ & $77.41_{\text{( -8.07)}}$ & $47.27_{\text{( -1.40)}}$ \\ \midrule
Gemma2-2B & $85.94_{\text{(+5.87)}}$ & $9.23_{\text{( -14.43)}}$ & $10.74_{\text{( -18.81)}}$ & $38.36_{\text{(+17.07)}}$ & $62.05_{\text{(+11.01)}}$ & $91.25_{\text{(+0.16)}}$ & $58.32_{\text{( -24.13)}}$ & $76.64_{\text{( -83.35)}}$ \\
Gemma2-9B & $93.34_{\text{(+3.24)}}$ & $13.77_{\text{( -9.80)}}$ & $14.75_{\text{( -11.40)}}$ & $34.82_{\text{(+10.87)}}$ & $61.50_{\text{(+7.77)}}$ & $91.73_{\text{( -0.13)}}$ & $61.24_{\text{( -17.70)}}$ & $61.94_{\text{( -26.80)}}$ \\
Gemma2-27B & $93.72_{\text{(+5.24)}}$ & $16.78_{\text{( -11.45)}}$ & $17.91_{\text{( -14.00)}}$ & $31.81_{\text{(+9.67)}}$ & $57.35_{\text{(+6.00)}}$ & $92.35_{\text{(+0.20)}}$ & $65.62_{\text{( -15.22)}}$ & $55.07_{\text{( -39.20)}}$ \\ \midrule
Phi3-mini & $78.87_{\text{( -0.98)}}$ & $41.68_{\text{( -7.86)}}$ & $52.85_{\text{( -9.19)}}$ & $\textbf{13.59}_{\text{(+2.87)}}$ & $\underline{43.34}_{\text{(+1.64)}}$ & $93.44_{\text{( -0.92)}}$ & $94.31_{\text{( -3.22)}}$ & $90.59_{\text{( -10.11)}}$ \\
Phi3-small & $90.64_{\text{(+0.85)}}$ & $42.05_{\text{( -7.87)}}$ & $46.40_{\text{( -9.21)}}$ & $\underline{14.27}_{\text{(+3.28)}}$ & $\textbf{42.06}_{\text{(\textbf{+1.25})}}$ & $\textbf{94.22}_{\text{(+0.00)}}$ & $94.75_{\text{( -2.26)}}$ & $58.47_{\text{( -18.55)}}$ \\
Phi3-medium & $92.24_{\text{(\underline{+6.40})}}$ & $47.83_{\text{( -2.38)}}$ & $51.85_{\text{( -6.64)}}$ & $16.21_{\text{(+5.31)}}$ & $45.43_{\text{(+5.74)}}$ & $93.71_{\text{( -1.45)}}$ & $92.63_{\text{( -5.60)}}$ & $85.77_{\text{(+5.15)}}$ \\ \midrule
Mistral-7B & $86.75_{\text{(+5.49)}}$ & $22.12_{\text{( -18.59)}}$ & $25.50_{\text{( -24.60)}}$ & $31.88_{\text{(+13.75)}}$ & $61.74_{\text{(+12.94)}}$ & $91.69_{\text{( -0.87)}}$ & $66.29_{\text{( -20.20)}}$ & $93.86_{\text{( -145.09)}}$ \\
Mistral-small & $97.24_{\text{(+1.89)}}$ & $29.85_{\text{( -7.27)}}$ & $30.70_{\text{( -8.23)}}$ & $30.94_{\text{(+8.01)}}$ & $62.80_{\text{(+7.25)}}$ & $90.70_{\text{( -0.93)}}$ & $64.38_{\text{( -14.01)}}$ & $109.91_{\text{(+12.03)}}$ \\
Mistral-large & $91.58_{\text{(+4.52)}}$ & $14.48_{\text{( -9.19)}}$ & $15.81_{\text{( -11.37)}}$ & $41.04_{\text{(+15.77)}}$ & $67.42_{\text{(+12.81)}}$ & $90.81_{\text{( -1.77)}}$ & $48.60_{\text{( -27.94)}}$ & $74.12_{\text{( -20.79)}}$ \\ \midrule
Mixtral-8x7B & $83.64_{\text{(+6.28)}}$ & $14.44_{\text{( -5.23)}}$ & $17.26_{\text{( -8.17)}}$ & $16.15_{\text{(+4.87)}}$ & $43.52_{\text{(+5.02)}}$ & $93.97_{\text{( -1.09)}}$ & $93.61_{\text{( -5.21)}}$ & $48.99_{\text{( -3.36)}}$ \\
Mixtral-8x22B & $91.59_{\text{(+0.63)}}$ & $21.43_{\text{( -14.74)}}$ & $23.40_{\text{( -16.37)}}$ & $20.96_{\text{(+7.56)}}$ & $49.03_{\text{(+6.69)}}$ & $90.55_{\text{( -4.00)}}$ & $81.77_{\text{( -14.31)}}$ & $\underline{45.93}_{\text{( -10.58)}}$ \\ \midrule
Qwen2.5-0.5B & $64.51_{\text{(+0.01)}}$ & $29.23_{\text{(+1.11)}}$ & $45.31_{\text{(+1.71)}}$ & $15.75_{\text{(+4.89)}}$ & $47.91_{\text{(+7.53)}}$ & $90.89_{\text{( -2.38)}}$ & $89.56_{\text{( -7.56)}}$ & $175.63_{\text{(+81.41)}}$ \\
Qwen2.5-1.5B & $62.76_{\text{(+5.67)}}$ & $10.52_{\text{( -4.14)}}$ & $16.76_{\text{( -8.93)}}$ & $25.50_{\text{(+10.86)}}$ & $55.05_{\text{(+11.83)}}$ & $92.16_{\text{( -2.12)}}$ & $76.82_{\text{( -14.89)}}$ & $72.33_{\text{(+7.51)}}$ \\
Qwen2.5-3B & $84.05_{\text{( -1.96)}}$ & $12.41_{\text{( -10.19)}}$ & $14.77_{\text{( -11.51)}}$ & $24.93_{\text{(+10.08)}}$ & $55.18_{\text{(+7.68)}}$ & $91.61_{\text{( -0.33)}}$ & $80.78_{\text{( -11.97)}}$ & $82.97_{\text{( -32.38)}}$ \\
Qwen2.5-7B & $90.08_{\text{(+2.27)}}$ & $17.61_{\text{( -10.97)}}$ & $19.55_{\text{( -13.00)}}$ & $22.23_{\text{(+7.04)}}$ & $52.50_{\text{(+6.53)}}$ & $92.64_{\text{( -0.30)}}$ & $83.96_{\text{( -8.39)}}$ & $63.78_{\text{( -12.19)}}$ \\
Qwen2.5-14B & $95.31_{\text{(+0.82)}}$ & $24.92_{\text{( -15.73)}}$ & $26.14_{\text{( -16.88)}}$ & $20.56_{\text{(+6.60)}}$ & $51.17_{\text{(+6.79)}}$ & $92.71_{\text{( -0.39)}}$ & $86.16_{\text{( -7.05)}}$ & $56.11_{\text{( -12.14)}}$ \\
Qwen2.5-32B & $97.12_{\text{(+0.63)}}$ & $40.02_{\text{( -7.20)}}$ & $41.21_{\text{( -7.73)}}$ & $14.61_{\text{(\underline{+1.67})}}$ & $44.89_{\text{(\underline{+1.33})}}$ & $92.15_{\text{( -0.65)}}$ & $92.47_{\text{( -0.73)}}$ & $72.87_{\text{(+1.84)}}$ \\
Qwen2.5-72B & $97.63_{\text{(+0.46)}}$ & $39.70_{\text{( -10.57)}}$ & $40.66_{\text{( -11.07)}}$ & $22.44_{\text{(+5.20)}}$ & $51.62_{\text{(+3.64)}}$ & $93.32_{\text{(+0.13)}}$ & $82.47_{\text{( -6.03)}}$ & $57.46_{\text{( -14.56)}}$ \\ \midrule
OLMo2-7B & $93.32_{\text{(+2.22)}}$ & $16.61_{\text{( -11.49)}}$ & $17.80_{\text{( -13.04)}}$ & $25.95_{\text{(+6.89)}}$ & $56.29_{\text{(+3.85)}}$ & $91.84_{\text{(+0.19)}}$ & $76.18_{\text{( -9.07)}}$ & $75.27_{\text{( -37.66)}}$ \\
OLMo2-13B & $95.74_{\text{(+0.52)}}$ & $18.11_{\text{( -9.08)}}$ & $18.92_{\text{( -9.64)}}$ & $31.63_{\text{(+7.30)}}$ & $61.68_{\text{(+3.63)}}$ & $90.48_{\text{(+0.47)}}$ & $66.07_{\text{( -11.13)}}$ & $117.15_{\text{(\textbf{ -316.35})}}$ \\ \midrule
T\"ulu3-8B & $94.20_{\text{(+4.41)}}$ & $16.68_{\text{( -10.88)}}$ & $17.70_{\text{( -12.98)}}$ & $26.23_{\text{(+10.13)}}$ & $54.80_{\text{(+8.82)}}$ & $92.41_{\text{( -0.20)}}$ & $76.87_{\text{( -13.55)}}$ & $55.90_{\text{( -47.23)}}$ \\
T\"ulu3-70B & $96.43_{\text{(+2.96)}}$ & $30.09_{\text{( -3.38)}}$ & $31.20_{\text{( -4.60)}}$ & $27.98_{\text{(+6.43)}}$ & $55.32_{\text{(+3.07)}}$ & $92.59_{\text{(+0.36)}}$ & $71.97_{\text{( -8.07)}}$ & $51.83_{\text{( -27.95)}}$ \\ \midrule
GPT-3.5 & $96.03_{\text{(+3.28)}}$ & $24.31_{\text{( -9.83)}}$ & $25.32_{\text{( -11.50)}}$ & $35.70_{\text{(+12.14)}}$ & $65.28_{\text{(+8.45)}}$ & $91.41_{\text{(\underline{+0.53})}}$ & $59.46_{\text{( -16.70)}}$ & $71.05_{\text{(\underline{ -165.03})}}$ \\
GPT-4o & $98.55_{\text{(+1.85)}}$ & $55.95_{\text{(+2.61)}}$ & $56.78_{\text{(+1.62)}}$ & $23.09_{\text{(+3.14)}}$ & $53.50_{\text{(+2.56)}}$ & $92.76_{\text{( -0.15)}}$ & $83.03_{\text{( -3.48)}}$ & $52.32_{\text{( -5.83)}}$ \\ \midrule
Gemini-Flash & $94.40_{\text{(+2.61)}}$ & $20.61_{\text{( -12.68)}}$ & $21.84_{\text{( -14.44)}}$ & $35.37_{\text{(+11.34)}}$ & $62.74_{\text{(+6.58)}}$ & $91.80_{\text{(+0.42)}}$ & $57.77_{\text{( -18.16)}}$ & $55.39_{\text{( -37.37)}}$ \\
Gemini-Pro & $93.23_{\text{(+4.08)}}$ & $35.52_{\text{(+2.84)}}$ & $38.10_{\text{(+1.44)}}$ & $20.20_{\text{(\textbf{+1.53})}}$ & $52.79_{\text{(+1.91)}}$ & $92.14_{\text{(\textbf{+0.64})}}$ & $84.23_{\text{( -0.25)}}$ & $62.57_{\text{( -54.52)}}$ \\

\bottomrule
\end{tabular}
}
\caption{Evaluation results for the generated sentences of each LLM with the \textbf{ideal example} one-shot setting. The average scores for Ordered CommonGen, where the order of concepts is specified, are reported across all six templates as the main results, with performance differences from the zero-shot Ordered CommonGen, as reported in Table~\ref{tab:main_results}, shown in parentheses. Bold scores indicate the LLM with the best performance for each metric, while underlined scores represent the second-best. Higher scores signify better performance for Concepts Coverage and Corpus-level Diversity, whereas lower scores indicate better performance for Sentence-level Similarity and Perplexity. Note that all metrics except perplexity are expressed as percentages. The values in parentheses indicate relative relationships between scores, showing increases or decreases as well as improvements or deteriorations based on the score differences. In other words, they are calculated using simple subtraction.}
\label{tab:1shot_results}
\end{table*}